\pdfoutput=1
\documentclass[11pt,a4paper]{article}
\usepackage[hyperref]{acl2020}
\usepackage{times}
\usepackage{latexsym}
\usepackage{amsmath}
\usepackage{amsfonts}
\usepackage{comment}

\usepackage{microtype}

\usepackage{multirow}
\usepackage{cleveref}
\crefformat{section}{\S#2#1#3}
\crefformat{subsection}{\S#2#1#3}
\crefformat{subsubsection}{\S#2#1#3}
\crefformat{appendix}{Appendix #2#1#3}
\crefformat{equation}{Eq.~(#2#1#3)}
\crefformat{table}{Table~#2#1#3}
\crefformat{figure}{Figure~#2#1#3}
\usepackage{booktabs}
\usepackage{enumitem}
\usepackage{graphicx}
\usepackage{subcaption}
\aclfinalcopy % Uncomment this line for the final submission
\def\aclpaperid{1716} %  Enter the acl Paper ID here

\newcommand{\squeezeup}{\vspace{-2.5mm}}

\title{Revisiting Unsupervised Relation Extraction}

\author{
Thy Thy Tran\textsuperscript{\textnormal{1}},
  Phong Le\textsuperscript{\textnormal{1}},
  Sophia Ananiadou\textsuperscript{\textnormal{1,2}} \\
  \textsuperscript{\textnormal{1}}National Centre for Text Mining, University of Manchester, United Kingdom 
  \\
  \textsuperscript{\textnormal{2}}The Alan Turing Institute, London, United Kingdom 
  \\
  \texttt{\{thy.tran, phong.le,  sophia.ananiadou\}@manchester.ac.uk} \\}

\date{}

\begin{document}
\maketitle
\begin{abstract}

Unsupervised relation extraction (URE) extracts relations between named entities from raw text without manually-labelled data and existing knowledge bases (KBs). URE methods can be categorised into generative and discriminative approaches, which rely either on hand-crafted features or surface form. However, we demonstrate that by using only named entities to induce relation types, we can outperform existing methods on two popular datasets. We conduct a comparison and evaluation of our findings with other URE techniques, to ascertain the important features in URE. We conclude that entity types provide a strong inductive bias for URE.\footnote{Source code is available at \url{https://github.com/ttthy/ure}}
\end{abstract}

\section{Introduction}
\label{sec:intro}

Relation extraction (RE) extracts semantic relations between entities from plain text.
For instance, ``\textbf{Jon Robin Baitz}$_{head}$ , born in \textbf{Los Angeles}$_{tail}$ ...''
expresses the relation \emph{/people/person/place\_of\_birth} between 
the two head-tail entities.
Extracted relations are then used for several downstream tasks such as 
information retrieval~\cite{corcoglioniti2016knowledge} and knowledge base construction~\cite{al2018extracting}.
RE has been widely studied using fully supervised 
learning~\cite{nguyen-grishman-2015-relation,miwa-bansal-2016-end,zhang-etal-2017-position,zhang-etal-2018-graph} 
and distantly supervised approaches~\cite{mintz-etal-2009-distant,riedel2010modeling,lin-etal-2016-neural}.

Unsupervised relation extraction (URE) methods have not been explored as much as fully or distantly supervised learning techniques.
URE is promising, since it does not require manually annotated data nor human curated knowledge bases (KBs), which are expensive to produce. 
Therefore, it can be applied to domains and languages where annotated data and KBs are not available. 
Moreover, URE can discover new relation types, since it is not restricted to specific relation types in the same way as fully and distantly supervised methods.
One might argue that Open Information Extraction (OpenIE) can also discover new relations.
However, OpenIE identifies relations based on textual surface information. Thus, similar relations with different textual forms may not be recognised. 
Unlike OpenIE, URE groups similar relations into clusters.
Despite these advantages, there are only a few attempts tackling 
URE using machine learning (ML) \cite{hasegawa-etal-2004-discovering, banko2007open, yao-etal-2011-structured,marcheggiani-titov-2016-discrete, simon-etal-2019-unsupervised}.

Similarly to other unsupervised learning tasks, a challenge in URE is how to evaluate results. 
Recent approaches~\cite{yao-etal-2011-structured,marcheggiani-titov-2016-discrete,simon-etal-2019-unsupervised} 
employ a widely used data generation setting in distantly supervised RE, i.e., aligning a large amount of raw text against triplets in a curated KB.
A standard metric score is computed by comparing the output relation clusters against the automatically annotated relations.
In particular, the \mbox{NYT-FB} dataset~\cite{marcheggiani-titov-2016-discrete} which is used for evaluation,
has been created by mapping relation triplets in Freebase~\cite{bollacker2008freebase} 
against plain text articles in the New York Times (NYT) corpus~\cite{sandhaus2008new}.
Standard clustering evaluation metrics for URE include B$^3$~\cite{bagga1998algorithms}, V-measure~\cite{rosenberg-hirschberg-2007-v}, and ARI~\cite{hubert1985comparing}.

Although the above mentioned experimental setting can be created automatically, there are three challenges to overcome.
Firstly, the development and test sets are silver, i.e., they include noisy labelled instances, since they are not human-curated.
Secondly, the development and test sentences are part of the training set, i.e., a transductive setting. 
It is thus unclear how well the existing models perform on unseen sentences. 
Finally, \mbox{NYT-FB} can be considered highly imbalanced, since only 2.1\% of the training sentences can be aligned with Freebase's triplets. 
Due to the noisy nature of silver data (\mbox{NYT-FB}), evaluation on silver data will not accurately reflect the system performance.
We also need unseen data during testing to examine the system generalisation.
To overcome these challenges, we will employ the test set of \mbox{TACRED}~\cite{zhang-etal-2017-position},
a widely used manually annotated corpus. 
Regarding the imbalanced data, we will demonstrate that in fact around 60\% (instead of 2.1\%) of instances in the training set express relation types defined in Freebase. 

In this work, we present a simple URE approach relying only on entity types that can obtain improved performance compared to current methods.
Specifically, given a sentence consisting of two entities and their corresponding entity types, e.g., PERSON and LOCATION, 
we induce relations as the combination of entity types, e.g., PERSON-LOCATION.
It should be noted that we employ only entity types because their combinations form reasonably coarse relation types 
(e.g., PERSON-LOCATION covers \emph{/people/person/place\_of\_birth} defined in Freebase).
We further discuss our improved performance in \cref{sec:discussion}.

Our contributions are as follows: (i) We perform experiments on both automatically/manually-labelled datasets, namely \mbox{NYT-FB} and TACRED, respectively. We show that two methods using only entity types can outperform the state-of-the-art models including both feature-engineering and deep learning approaches.  The surprising results raise questions about the current state of unsupervised relation extraction. (ii) For model design, we show that link predictor provides a good signal to train a URE model (Fig 1). We also illustrate that entity types are a strong inductive bias for URE (\cref{tab:results}).

\section{Methods for URE}
The goal of URE is to predict the relation $r$ between two entities $e_{head}$ and $e_{tail}$ in a sentence $s$.
We will describe three recent ML-based methods tackling URE and our own methods.
We divide the ML-based methods into two main approaches: generative and discriminative.

\subsection{Generative Approach}
\citet{yao-etal-2011-structured} 
extended topic modelling -- Latent Dirichlet Allocation (LDA)~\cite{blei2003latent} for RE, developing two models, herewith \textbf{RelLDA} and \textbf{RelLDA1}.
In both models, a sentence and an entity pair perform as a document in topic modelling, while a relation type corresponds to a topic.
RelLDA uses three features, i.e., the shortest dependency path between two entities and the two entity mentions.
RelLDA1 extends RelLDA with five more features, i.e., the entity types, words and part-of-speech tags between the two entities. 

\subsection{Discriminative Approaches}
\citet{marcheggiani-titov-2016-discrete} proposed a discrete-state variational autoencoder (VAE) to tackle URE (herewith \textbf{March}).
Their model consists of two components: a \emph{relation classifier} and a \emph{link predictor}. 
The \emph{relation classifier}, which is discriminative, takes entity types and several linguistic features (e.g., dependencies) as input to predict the relation $r$. 
The \emph{link predictor} then uses the (soft) predicted relation $r$  to predict the missing entity $e_i$ in a specific position 
$\{\text{head}, \text{tail}\}$, given the other entity $e_{-i}$, 
where if $i=\text{head}$ then $-i=\text{tail}$ and vice versa.
In other words, entity prediction, in a self-supervised manner, provides training signals to learn the relation classifier. 
However, by using only entity prediction, only a few relation types are chosen. 
They thus used \emph{entropy} over all relations as a regulariser.
The maximisation of the \emph{entropy} regulariser ensures the uniform relation distribution and allows more relations to be predicted.

Another discriminative method is by \citet{simon-etal-2019-unsupervised} (herewith \textbf{Simon}) which differs from March in the following ways: 
a) firstly, its relation classifier employs a piece-wise convolutional network (PCNN) using only surface form without requiring hand-crafted features; 
b) secondly, they replaced \emph{entropy} with two regularisers: 
$L_s$ (\emph{skewness}), to encourage the relation classifier to be confident in its prediction, and $L_d$ (\emph{dispersion}), to ensure several relation types are predicted over a minibatch. 
Note that, $L_s$ is equivalent to the negation of the \emph{entropy} used in March. 

\subsection{Our Methods}
We introduce two entity-based methods, herewith \textbf{EType} and \textbf{EType+}.
Our motivation is that entity types are helpful for RE, as mentioned in \newcite{zhang-etal-2017-position} for supervised learning and \newcite{ren2017cotype} for distant learning. 
In URE, \newcite{yao-etal-2011-structured, marcheggiani-titov-2016-discrete} also used entity types.
We therefore propose EType that induces coarse relation clusters from the entity types.
In particular, given two entity types $t_{e_{head}}$, $t_{e_{tail}}$ as input, EType would output their concatenation $t_{e_{head}}$-$t_{e_{tail}}$ as the relation.

One problem with EType is that the number of relation types is  determined by the number of entity types. 
For instance, 4 entity types lead to $4^2=16$ relation types. 
To extract an arbitrary number of relation types, we build a relation classifier that consists of one-layer feed-forward network taking entity type combinations as input: 
\[
r=\mathbf{FFN}(t_{e_{head}}\text{-}t_{e_{tail}}),\]
where $t_{e_{head}}$-$t_{e_{tail}}$ is the one hot vector of the entity type pair. 
We then employ the link predictor used in March and the two regularisers used in Simon, to produce a new method, herewith EType+.

\section{Experiments and Results}
\label{sec:exps}

\paragraph{Evaluation metrics}
We use the following evaluation metrics for our analysis: a) B$^3$~\cite{bagga1998algorithms} used in previous work, which is the harmonic mean of precision and recall for clustering task; 
b) V-measure~\cite{rosenberg-hirschberg-2007-v},
and c) ARI~\cite{hubert1985comparing} used in \citet{simon-etal-2019-unsupervised}.~\footnote{We used sklearn.metrics package to compute V-measure and ARI.}
V-measure is analysed in terms of homogeneity and completeness, while ARI measures the similarity between two clusterings.
We note that V-measure is sensitive to the dependency between the number of clusters and instances.
A relatively small number of clusters compared to the number of instances should be used to maintain the comparability of using V-measure.
More precisely, we evaluated V-measure of the trivial homogeneity, where there are only singular clusters (i.e., each instance is its own cluster).
The V-measure of the trivial homogeneity on \mbox{NYT-FB} reached 43.77\%, which is higher than all the implemented methods in \cref{tab:results}. 
Meanwhile, neither B$^3$ nor ARI encounters this problem.
\squeezeup
\paragraph{Datasets} 
We employed \mbox{NYT-FB} for training and evaluation following previous work~\cite{yao-etal-2011-structured,marcheggiani-titov-2016-discrete, simon-etal-2019-unsupervised}.
Because only 2.1\% of the sentences in \mbox{NYT-FB} were aligned against Freebase's triplets, 
we were concerned whether this dataset contains enough sentences for a model to learn relation types from Freebase. 
We thus examined $100$ randomly chosen instances from 1.86m non-aligned sentences. 
We found that 61\% of them (or 60\% of the whole dataset) express relation types in Freebase. 
This suggests that the \mbox{NYT-FB} dataset can be employed to train a relation extractor.
However, there are two further issues when evaluating URE methods on \mbox{NYT-FB}.
Firstly, the development and test sets are all aligned sentences without human curation, which means that they include wrong/noisy labelled instances. 
In particular, we found that 35 out of 100 randomly chosen sentences were given incorrect relations.
Secondly, the two validation sets are part of the training set. This setting is obviously not inductive, as it does not evaluate how a model performs on unseen sentences. 
Therefore, we \emph{additionally evaluate} all methods (except topic modelling) on the test set of \mbox{TACRED}~\cite{zhang-etal-2017-position}, a widely used manually annotated corpus for supervised RE.
The statistics of both \mbox{NYT-FB} and \mbox{TACRED} are provided in~\cref{secapp:datasets}.
\squeezeup

% ===============================
\begin{table}[t!]
\small 
\centering
\begin{tabular}{lcccc}
\toprule
\multicolumn{2}{l}{\textbf{Model}} & \textbf{B\textsuperscript{3}} & \textbf{V} & \textbf{ARI} \\ \midrule
\multicolumn{5}{c}{\mbox{NYT-FB}} \\ \midrule
RelLDA & \multirow{5}{*}{$c=10$} & 29.1 & 30.0 & 13.3 \\
RelLDA1 &  & 36.9 & 34.7 & 24.2 \\
March ($L_s$+$L_d$)  && 37.5  & 38.7  & 27.6 \\
Simon   && 39.4  & 38.3  & \textbf{33.8} \\
EType+  && \textbf{41.9} & 40.6 & 30.7 \\ 
\cmidrule{2-5}
March\textsuperscript{$\diamond$} ($L_s$+$L_d$)  && 36.9 & 37.4 & 28.1 \\
EType   & \multirow{2}{*}{$c=16$} & 41.7 & \textbf{42.1} & 30.7 \\
EType+  && 41.5 & 41.3 & 30.5 \\
\cmidrule{2-5}
RelLDA1 & \multirow{2}{*}{$c=100$} & 29.6 & - & - \\
March   && 35.8 & - & - \\ 
\midrule
\multicolumn{5}{c}{\mbox{TACRED}}\\ \midrule
March\textsuperscript{$\diamond$} ($L_s$+$L_d$)  & \multirow{3}{*}{$c=10$} &  31.0 & 43.8 & 22.6\\
Simon\textsuperscript{$\diamond$}  && 15.7 & 17.1 & 6.1 \\
EType+ && 43.3 & 59.7 & 25.7\\
\cmidrule{2-5}
March\textsuperscript{$\diamond$} ($L_s$+$L_d$)  & \multirow{3}{*}{$c=16$} & 34.6 & 47.6 & 23.2 \\
EType  && \textbf{48.3} & \textbf{64.4} & \textbf{29.1} \\
EType+ && 46.1 & 62.0 & 27.4 \\
\cmidrule{2-5}
March\textsuperscript{$\diamond$}  & $c=100$ &33.13 & 43.63& 20.21 \\
\bottomrule
\end{tabular}
\caption{\label{tab:results} Average results (\%) across three runs of different models (except the  EType) on 
\mbox{NYT-FB} and \mbox{TACRED}. $c$ indicates the number of clusters in each method. \textsuperscript{$\diamond$} indicates our implementation of the corresponding model.
We note that all methods were trained on \mbox{NYT-FB} and evaluated on the test set of both \mbox{NYT-FB} and \mbox{TACRED}.
}
\end{table}

% =========================

\paragraph{Hyper-parameters}
We examine three models RelLDA1, March, and Simon using the reported hyper-parameters~\cite{yao-etal-2011-structured,marcheggiani-titov-2016-discrete, simon-etal-2019-unsupervised}. 
For comparison, we also evaluate March with the two regularisers of Simon, namely \textbf{March ($L_s+L_d$)}.
To evaluate on \mbox{TACRED}, we employed the original March with $n=100$ using the published repository\footnote{\href{https://github.com/diegma/relation-autoencoder}{github.com/diegma/relation-autoencoder}}.
Meanwhile, for March ($L_s$+$L_d$) and Simon, we reimplemented these models and evaluated them on \mbox{TACRED}.
Regarding our methods, EType does not have hyper-parameters, while EType+ uses the same optimiser and entity type dimension as in Simon.
All the hyper-parameters used in our experiments are listed in~\cref{secapp:hyper}. 

\squeezeup

\paragraph{Results} \cref{tab:results} demonstrates the average performance of our methods across three runs in comparison with the three ML models on \mbox{NYT-FB} and \mbox{TACRED}.
Our models outperform the best performing system of~\newcite{simon-etal-2019-unsupervised} on both datasets, except ARI on \mbox{NYT-FB}.
ARI is shown to be used when there are large equal-sized clusters~\cite{romano2016adjusting} while relation datasets are generally imbalanced (both \mbox{NYT-FB} and \mbox{TACRED} in this study; please refer to~\cref{secapp:datasets} for the detailed statistics). 
Due to this reason, ARI might not be appropriate to evaluate URE systems.
In addition, the ML methods consistently exhibit lower performance on \mbox{TACRED} than on \mbox{NYT-FB}. 
The full results are shown in~\cref{secapp:results}.

\label{sec:discussion}
\begin{figure}[t!]
    \centering
    \includegraphics[width=0.9\linewidth]{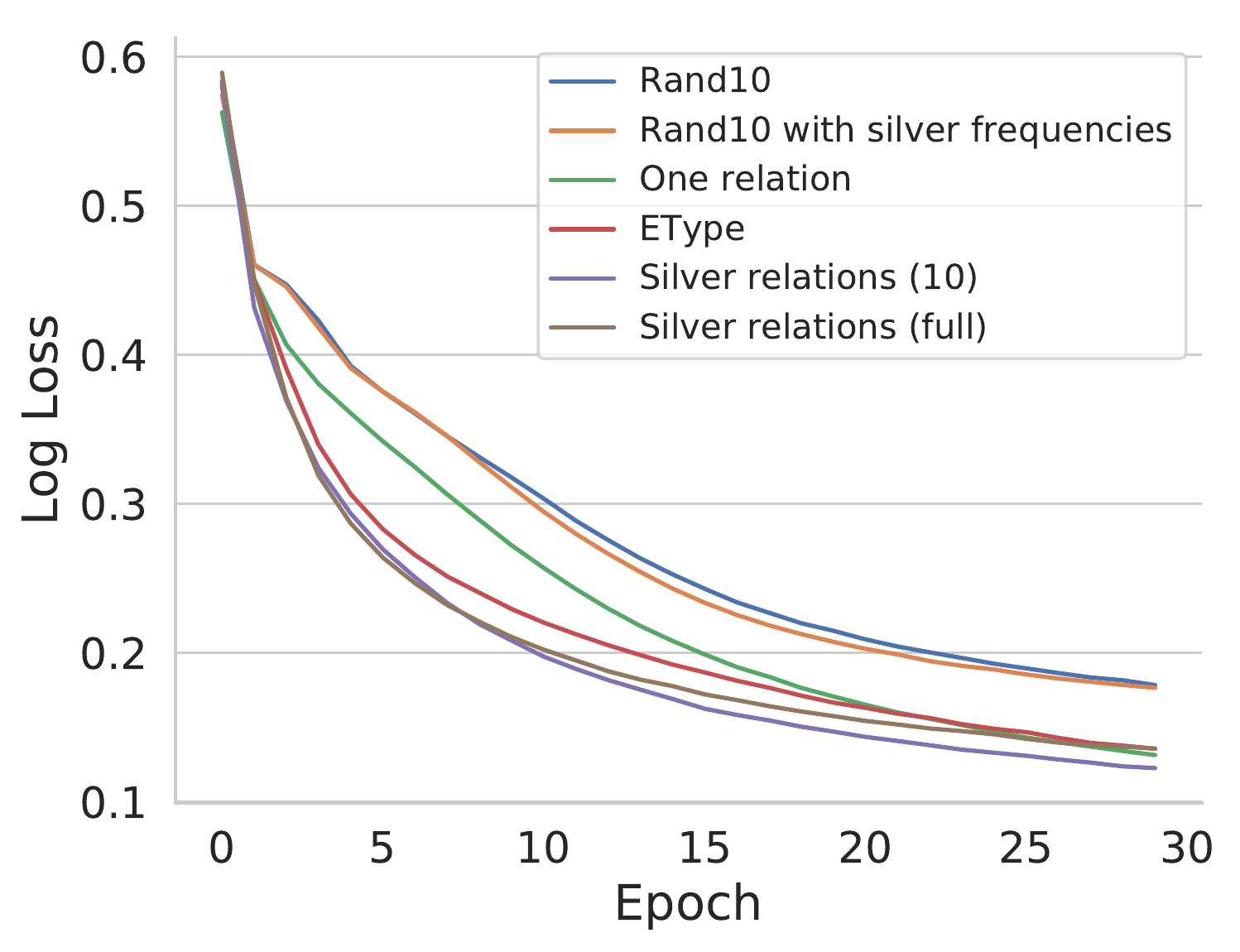}
    \caption{Average negative log likelihood losses across three runs of the link predictor on the training data (not including negative instances). Each line demonstrates a different relation input setting.
    }
    \label{fig:lp}
\end{figure}

\section{Discussion}

The results of our evaluation demonstrate that our models outperform previous methods, despite being simpler than them. These results lead us to the following findings.

\paragraph{Do ML models employ proper inductive biases?} In common with other unsupervised learning approaches, there is no guarantee that a URE model would learn the relation types in the used KBs and/or annotated data. 
A common solution is to employ inductive biases~\cite{wagstaff2000refining} to guide the learning process towards desired relation types. 
Inductive biases can emanate from pre-processed data. 
Since our models outperform other methods, we conclude that entity type information alone constitutes a better bias than the biases employed by existing ML models. 
Indeed, entity types constitute a useful bias for this task. 
Among the topic modelling based methods, RelLDA1 outperforms RelLDA, which does not employ entity types. 
In a separate  experiment, we found that adding entity types to the Simon model helped to achieve higher performance than the original version, i.e., 42.74\% vs. 39.4\% F1 B$^3$ on the \mbox{NYT-FB} test set.
However, although both RelLDA1 and March also employ entity types, their performance is still lower than ours. 
This is because other syntactic and word features used in these two models might cancel out the useful bias of entity types.
(More details are in the last paragraph of this section.)

Inductive biases can emanate from training signals. 
March and Simon are trained from a link predictor, which provides indirect signals to train a relation classifier. 
Hence, the question here is \emph{``can the link predictor induce good training signals?''} 
To answer this, we examine the link predictor with alternative settings:
\begin{itemize}[noitemsep,nosep]
    \item \textbf{Rand10} randomly assigns one among 10 relation types to each entity pair; 
    \item \textbf{Rand10 with silver frequencies}, similar to \emph{Rand10}, randomly generates relation types but follows the silver relation distribution;
    \item \textbf{One relation} assumes all entity pairs sharing the same relation type;
    \item \textbf{EType} uses 16 relation types induced from 4 coarse entity types;
    \item \textbf{Silver relations (10)} takes the top 9 most frequent relation types and groups the rest together to form the tenth relation type; 
    \item \textbf{Silver relations (full)} considers the full (silver) annotated relations, i.e., 262 types.
\end{itemize}
\cref{fig:lp} illustrates the average loss values of using these settings. 
If high quality relations were critical for training the link predictor, we would expect lower losses while using annotated relations. 
Indeed, the loss curve of using 10 correct relation types is consistently below all the others. 
This implies that the link predictor is able to provide reasonable signals for training a relation classifier. 
So why are the Simon and March models outperformed by our models? As pointed out by \newcite{simon-etal-2019-unsupervised}, 
the link predictor itself cannot be trained without a good relation classifier. 
It suggests that the relation classifiers in both methods need to be improved. 
Empirical evidence shows that both Simon and March models are outperformed (in B$^3$ and V) 
by our Etype+, which uses the same link predictor.
We also notice that both \textit{One relation} and \textit{EType} at the end sharing similar performances.
This might imply that we only need one relation (matrix) to predict head/tail entities, as the link predictor is very expressive.
However, the silver relations are clearly helpful as during the first 15 epochs their losses are much lower than others.

\paragraph{Why was the performance on \mbox{TACRED} lower?}
Despite the fact that \mbox{TACRED} shares similar relation types with Freebase, we observed that both the March and Simon models consistently fare less well in terms of their performance on the \mbox{TACRED} dataset. 
More precisely, Simon model results in significantly worse performance on TACRED, with 15.7\% in terms of B\textsuperscript{3}, which is twice as low as on \mbox{NYT-FB} (39.4\%).
This performance drop might be attributed to the distributional shift of the two datasets: variation and semantic shift in vocabulary and language structure over time, since NYT was collected long before TACRED.

\begin{table}[t!]
\centering
\begin{tabular}{lllll}
\toprule
\multicolumn{2}{l}{\textbf{Model}} 
& \textbf{B\textsuperscript{3}} & \textbf{V} & \textbf{ARI} \\
\midrule
\multicolumn{2}{l}{EType+}      &       42.5    & 40.1  & 29.2       \\
  & +Entity & 40.5   & 39.9   & 28.6         \\
  & +BOW      & 37.7   & 38.0   & 20.5         \\
  & +DepPath      & 41.4   & 39.4   & 26.7         \\
  & +POS      & 41.6   & 40.4   & 27.8         \\
  & +Trigger  & 41.7   & 41.3   & 29.0        \\
  & +PCNN  & 40.8 &	39.6 &	27.1        \\
\bottomrule
\end{tabular}
\caption{\label{tab:combination}Study of \mbox{EType+} in combination with different features. The results are average across three runs on the development set.}
\end{table}

\paragraph{How is the performance when combining entity types with other features?} 
Our experiments using only entity types surprisingly perform higher than the previous state-of-the-art methods including feature engineering and deep learning models. However, we know that context information is crucial to distinguish the relation between two entities, as many RE studies have been proposed to integrate the context information to improve the RE performance. 
We conduct experiments when combining entity types with common features for RE in~\cref{tab:combination}.
The list of features include: (i) Entity: textual surface form of two entities, (ii) BOW: bag of words between two entities, (iii) DepPath: words on the dependency path between two entities, (iv) POS: part-of-speech tag sequence between two entities, and (v) Trigger: DepPath without stop words.
In general, naively combining entity types with other features could not improve the model performance. 
Additionally, BOW feature had negative effects on the RE performance. 
This indicates that bag of words between two entities often include uninformative and redundant words, i.e., noises, that are difficult to eliminate using simple neural architectures.
While (i)-(v) are widely used hand-crafted features for RE, we also incorporated a neural-based context encoder PCNN which is the combination of \emph{Simon}'s PCNN encoder, the entity masking and position-aware attention proposed in \cite{zhang-etal-2017-position}.
However, the performance of combining PCNN is also lower than only entity types.

\section{Conclusion}
\label{sec:conclusion}
We have shown the importance of entity types in URE.
Our methods use only entity types, yet they yield higher performance than previous work on both \mbox{NYT-FB} and TACRED. 
We have investigated the current experimental setting, concluding that a strong inductive bias is required to train a relation extraction model without labelled data.
URE remains challenging, which requires improved methods to deal with silver data.
We also plan to use different types of labelled data, e.g., domain specific data sets, to ascertain whether entity type information is more discriminative in sub-languages.

\section*{Acknowledgments}
We would like to thank the reviewers for their comments, Diego Marcheggiani for sharing his dataset with us, and \'Etienne Simon for sharing the hyperparameters. 
The first author thanks the University of Manchester for the Research Impact Scholarship Award.
This work is also funded by Lloyd’s Register Foundation, Discovering Safety Programme, Thomas Ashton Institute.

\bibliography{acl2020}
\bibliographystyle{acl_natbib}

\appendix

\section{Datasets}
\label{secapp:datasets}

\cref{tab:statistics} shows the statistics of the NYT-FB~\cite{marcheggiani-titov-2016-discrete} and TACRED~\cite{zhang-etal-2017-position} datasets.
We followed the same data split and pre-processing described in~\citet{marcheggiani-titov-2016-discrete}.
For all methods, we trained on \mbox{NYT-FB} and evaluated them on both \mbox{NYT-FB} and \mbox{TACRED}.

\begin{figure*}[t!]
     \centering
	\begin{subfigure}[t]{0.48\textwidth}
	\centering
     \includegraphics[width=\linewidth]{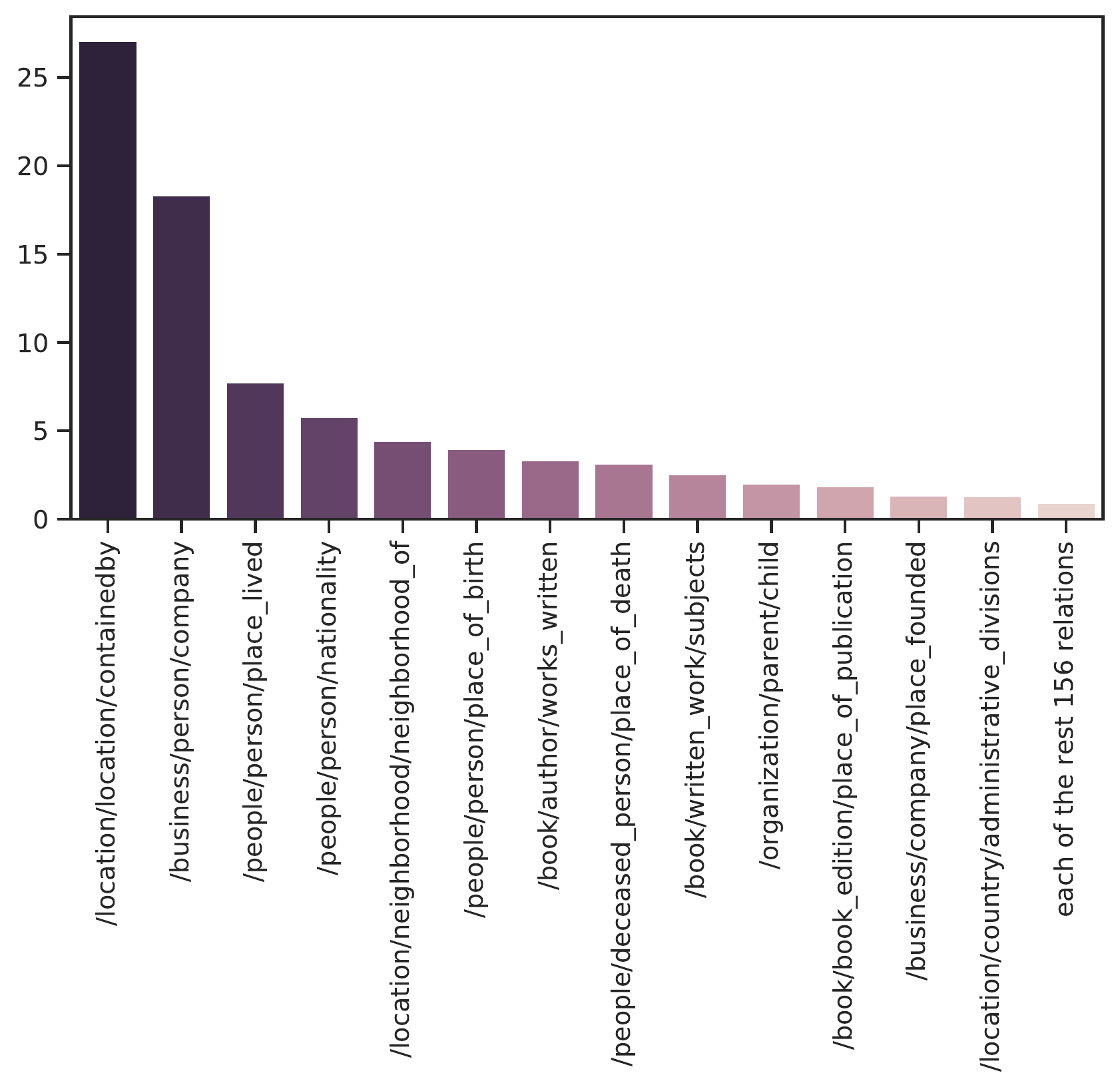}
    \caption{NYT-FB has 253 relation types in total \label{fig:nyt}}
    \end{subfigure}
	\quad
	\begin{subfigure}[t]{0.48\textwidth}
    	\centering
         \includegraphics[width=\linewidth]{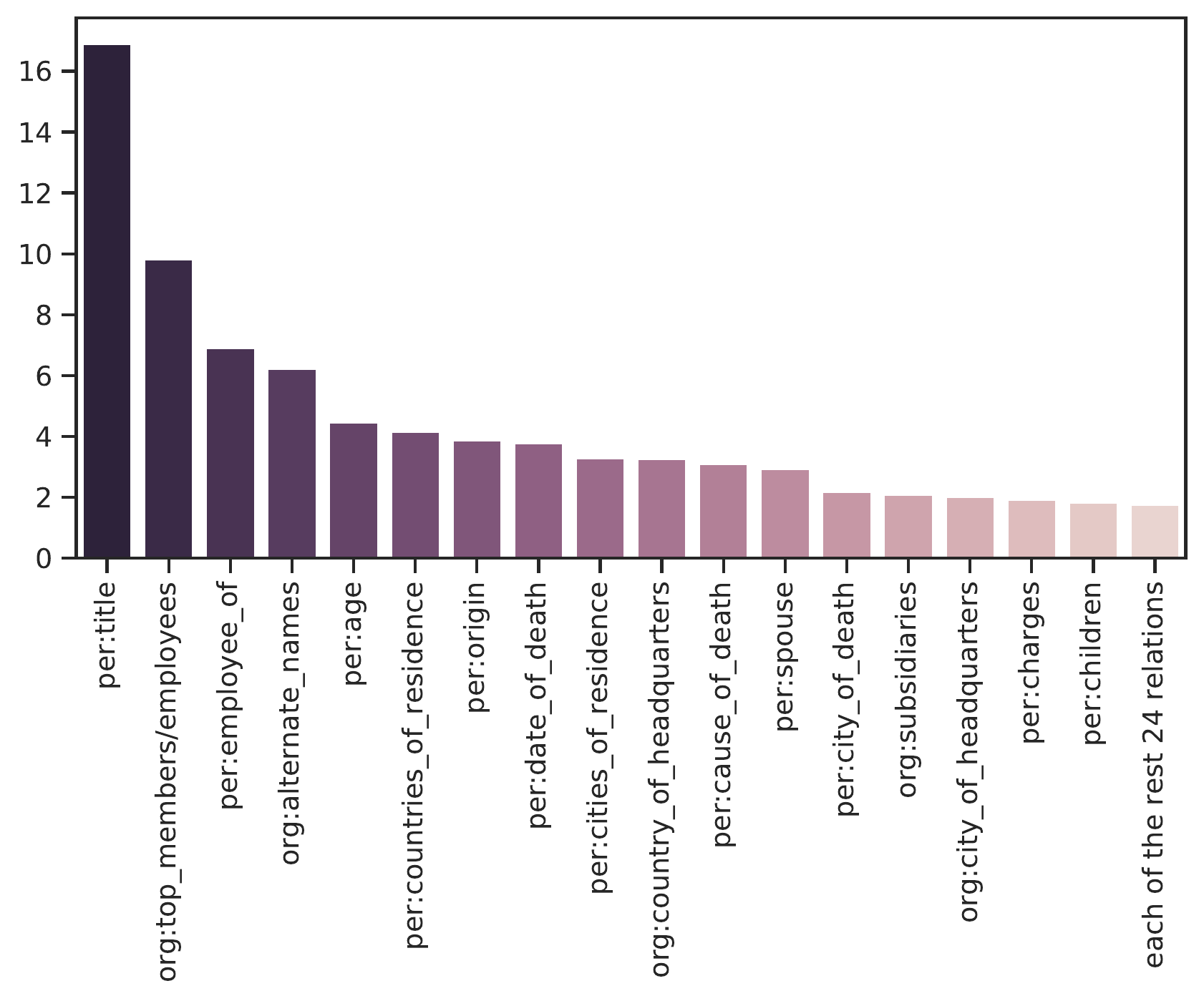}
        \caption{TACRED has 41 relation types in total\label{fig:tacred}}
     \end{subfigure}
    \caption{\label{fig:rel-stat} Relation distribution of \mbox{NYT-FB} and \mbox{TACRED} (\%).}
\end{figure*}

\cref{fig:rel-stat} illustrates the relation distributions of two datasets: \mbox{NYT-FB} and TACRED. We can see that 15/253 most frequent relations account for 82.97\% of the total number of instances in \mbox{NYT-FB}. Meanwhile, 
15/41 relations sum upto 74.94\% of the total number of instances in \mbox{TACRED}.

\section{Hyper-parameter Settings}
\label{secapp:hyper}

We used the development set to stop the training process.
For every model, we conducted three runs with different initialised parameters and computed the average performance.
We list the hyper-parameters of different models in \cref{tab:hyperparams}.

\section{Detailed Results}
\label{secapp:results}

\cref{tab:details} presents the average test scores of three runs on the \mbox{NYT-FB} and \mbox{TACRED} datasets.
We note that the two models proposed by \citet{marcheggiani-titov-2016-discrete} and \citet{simon-etal-2019-unsupervised} are sensitive to the hyper-parameters and thus difficult to train.
We could not replicate the performance of Simon on the \mbox{NYT-FB} dataset.

\begin{table}[th!]
\small
\centering
\begin{tabular}{lrrrr}
\toprule
& & \textbf{Train} & \textbf{Dev}     & \textbf{Test}      \\
            \midrule
\multicolumn{5}{c}{NYT-FB (\#$r=262$)}                                             \\
\midrule
\multicolumn{2}{l}{Raw instances} & 1,950,557 & 389,819 & 1,560,738 \\
            & Positive            & 41,685    & 7,793   & 33,808    \\
            \midrule
\multicolumn{5}{c}{\mbox{TACRED} (\#$r=41$)}                                          \\
\midrule
\multicolumn{2}{l}{Raw instances} & 68,124    & 22,631  & 15,509    \\
            & Positive            & 13,012    & 5,436   & 3,325    \\
            \bottomrule
\end{tabular}
\caption{\label{tab:statistics} The statistics of the \mbox{NYT-FB} and the \mbox{TACRED} datasets. 
\#$r$ indicates the number of relation types in each dataset.}
\end{table}

\begin{table}[th!]

\begin{subtable}[t]{0.45\textwidth}
\centering
\begin{tabular}{lrr}
\toprule
\textbf{Parameter}      & \textbf{$Ls$} & \textbf{$Ls+Ld$} \\
\midrule
Optimiser               &   \multicolumn{2}{c}{AdaGrad} \\
% python -u -m feature.reldist_main     --k_samples 5     --loss_coef_alpha 0.01     --loss_coef_beta 0.02     --lr 0.005     --n_epochs 50     --weight_decay 1e-7     --n_rels 16 
Number of epochs    & \multicolumn{2}{c}{10} \\
Batch size          & \multicolumn{2}{c}{100} \\
L2 regularisation   & \multicolumn{2}{c}{1e-7} \\
Feature dimension   & \multicolumn{2}{c}{10} \\
Learning rate       & 0.1 & 0.005 \\
$L_s$ coefficient  & 0.1  & 0.01 \\
$L_d$ coefficient  & --   & 0.02  \\
\bottomrule
\end{tabular}
\caption{\label{tab:march} \citet{marcheggiani-titov-2016-discrete}'s model.}
\end{subtable}
\vfill

\begin{subtable}[t]{0.45\textwidth}
\centering
\begin{tabular}{lr}
\toprule
\textbf{Parameter}      & \textbf{Value} \\
\midrule
Optimiser               &    Adam     \\
Learning rate           &    0.005  \\
Learning rate annealing &    $0.5^{0.25}$ \\
Batch size              &    100      \\
Early stop patience     &    10       \\
L2 regularisation       &    2e-11    \\
Word dimension          &      50     \\
Entity type dimension   &       10    \\
$L_s$ coefficient       &    0.01     \\
$L_d$ coefficient       &     0.02    \\
\bottomrule
\end{tabular}
\caption{\label{tab:simon}\citet{simon-etal-2019-unsupervised}'s model.}
\end{subtable}
\vfill

% python -u -m ure.etypeonly.main     --k_samples 5     --loss_coef_alpha 0.0001     --loss_coef_beta 0.02     --lr 0.001     --n_epochs 50     --weight_decay 1e-5     --n_rels 10     --patience 10 
\begin{subtable}[t]{0.45\textwidth}
\centering
\begin{tabular}{lr}
\toprule
\textbf{Parameter}      & \textbf{Value} \\
\midrule
Optimiser               &    Adam     \\
Learning rate           &    0.001  \\
Batch size              &    100      \\
Early stop patience     &    10       \\
L2 regularisation       &    1e-5    \\
% Word dimension          &      50     \\
Entity type dimension   &       10    \\
$L_s$ coefficient       &    0.0001     \\
$L_d$ coefficient       &     0.02    \\
\bottomrule
\end{tabular}
\caption{\label{tab:etype}EType+.}
\end{subtable}

\caption{\label{tab:hyperparams}Hyper-parameter values used in our experiments.}
\end{table}

\begin{table*}[t!]
\small
\centering
\begin{tabular}{lcccccccccc}
\toprule
\multicolumn{2}{l}{\multirow{2}{*}{\textbf{Model}}}                             & \multicolumn{3}{c}{\textbf{B\textsuperscript{3}}}                                                 &  & \multicolumn{3}{c}{\textbf{V-measure}}                                                                   &  & \multicolumn{1}{c}{\multirow{2}{*}{\textbf{ARI}}} \\ 
\multicolumn{2}{l}{}                                                            & \multicolumn{1}{c}{\textbf{F1}} & \multicolumn{1}{c}{\textbf{P}} & \multicolumn{1}{c}{\textbf{R}} &  & \multicolumn{1}{c}{\textbf{F1}} & \multicolumn{1}{c}{\textbf{Hom.}} & \multicolumn{1}{c}{\textbf{Comp.}} &  & \multicolumn{1}{c}{}                              \\\midrule
\multicolumn{11}{c}{NYT-FB} \\\midrule
RelLDA                            & \multicolumn{1}{c}{\multirow{7}{*}{$n=10$}} & 29.1                            & 24.8                           & 35.2                           &  & 30.0                            & 26.1                              & 35.1                               &  & 13.3                                              \\
RelLDA1                           & \multicolumn{1}{c}{}                        & 36.9                            & 30.4                           & 47.0                           &  & 37.4                            & 31.9                              & 45.1                               &  & 24.2                                              \\
March ($L_s$+$L_d$)               & \multicolumn{1}{c}{}                        & 37.5                            & 31.1                           & 47.4                           &  & 38.7                            & 32.6                              & 47.8                               &  & 27.6                                              \\
March ($L_s$+$L_d$)$\ddagger$ & \multicolumn{1}{c}{}                        & 38.7                            & 30.9                           & 51.7                           &  & 37.6                            & 31.0                              & 47.7                               &  & 26.1                                              \\
Simon                             & \multicolumn{1}{c}{}                        & 39.4                            & 32.2                           & 50.7                           &  & 38.3                            & 32.2                              & 47.2                               &  & 33.8                                              \\
Simon$\ddagger$               & \multicolumn{1}{c}{}                        &  32.6 & 28.2  & 38.9 &&       30.5 &   26.1     & 36.8 && 23.8           \\
EType+                            & \multicolumn{1}{c}{}                        & 41.9                            & 31.3                           & 63.7                           &  & 40.6                            & 31.8                              & 56.2                               &  & 30.7                                              \\\midrule
March ($L_s$+$L_d$)$\ddagger$ & \multirow{3}{*}{$n=16$}                     & 36.9                            & 32.0                           & 43.7                           &  & 37.4                            & 32.6                              & 43.9                               &  & 28.1                                              \\
EType                             &                                             & 41.7                            & 32.5                           & 58.0                           &  & 42.1                            & 34.7                              & 53.6                               &  & 30.7                                              \\
EType+                            &                                             & 41.5                            & 32.0                           & 59.0                           &  & 41.3                            & 33.6                              & 53.9                               &  & 30.5                                              \\\midrule
RelLDA1    & \multirow{3}{*}{$n=100$}  & 29.6 & -& - &  & -& -& -&  & -\\
March & & 35.8 & -& -&  & - & - & -&  & -\\
March$\ddagger$               &                                             & 34.8                            & 24.4                           & 62.4                           &  & 25.9                            & 18.7                              & 42.7                               &  & 13.1                                              \\\midrule
\multicolumn{11}{c}{TACRED}     \\\midrule
March ($L_s$+$L_d$)$\ddagger$ & \multirow{3}{*}{$n=10$}                     & 31.0                            & 21.7                           & 54.9                           &  & 43.8                            & 35.5                              & 57.2                               &  & 22.6                                              \\
Simon$\ddagger$  && 15.7 & 12.1 & 22.4 && 17.1 & 14.6 & 20.6 && 6.1 \\
EType+                            &                                             & 43.3                            & 28.0                           & 96.9                           &  & 59.7                            & 43.4                              & 96.0                               &  & 25.7                                              \\\midrule
March ($L_s$+$L_d$)$\ddagger$ & \multirow{3}{*}{$n=16$}                     & 34.6                            & 24.3                           & 61.3                           &  & 47.6                            & 38.9                              & 61.4                               &  & 23.2                                              \\
EType                             &                                             & 48.3                            & 32.3                           & 96.3                           &  & 64.4                            & 48.6                              & 95.6                               &  & 29.1                                              \\
EType+                            &                                             & 46.1                            & 30.3                           & 96.9                           &  & 62.0                            & 45.8                              & 96.1                               &  & 27.4                                              \\\midrule
March$\ddagger$               & $n=100$                                     & 33.13                           & 21.83                          & 69.20                          &  & 43.63                           & 32.96                             & 64.66                              &  & 20.21                                            \\ \bottomrule
\end{tabular}
\caption{\label{tab:details}Average results (\%) across three runs of different models (except the rule-based EType) on 
two datasets: the distant supervision \mbox{NYT-FB} and the large supervised dataset TACRED.  The model of \citet{marcheggiani-titov-2016-discrete} is March and the model of \citet{simon-etal-2019-unsupervised} is Simon. $\ddagger$ indicates our implementation of the corresponding model.}
\end{table*}

\end{document}